\documentclass[conference]{IEEEtran}
\IEEEoverridecommandlockouts
\usepackage{cite}
\usepackage{amsmath,amssymb,amsfonts}
\usepackage{algorithmic}
\usepackage{graphicx}
\usepackage{textcomp}
\usepackage{xcolor}
\usepackage[ruled,vlined]{algorithm2e}
\def\BibTeX{{\rm B\kern-.05em{\sc i\kern-.025em b}\kern-.08em
    T\kern-.1667em\lower.7ex\hbox{E}\kern-.125emX}}
\begin{document}

\makeatletter
\def\ps@IEEEtitlepagestyle{%
  \def\@oddfoot{\mycopyrightnotice}%
  \def\@evenfoot{}%
}
\def\mycopyrightnotice{%
  { \footnotesize 978-1-6654-4059-2/21/\$31.00 \copyright 2021 IEEE\hfill}
}
\makeatother
\title{Action Recognition using Transfer Learning and Majority Voting for CSGO \\
}

\author{\IEEEauthorblockN{ Tasnim Sakib Apon}
\IEEEauthorblockA{\textit{Computer Science and Engineering} \\
\textit{BRAC University}\\
Dhaka, Bangladesh \\
sakibapon7@gmail.com}
\and
\IEEEauthorblockN{Abrar Islam}
\IEEEauthorblockA{\textit{Electrical and Electronic Engineering} \\
\textit{Islamic University of Technology}\\
Dhaka, Bangladesh \\
abrarislam@iut-dhaka.edu}
\and
\IEEEauthorblockN{ MD. Golam Rabiul Alam}
\IEEEauthorblockA{\textit{dept. of Computer Science and Engineering} \\
\textit{BRAC University}\\
Dhaka, Bangladesh \\
rabiul.alam@bracu.ac.bd}

}

\maketitle

\begin{abstract}
Presently online video games have become a progressively favorite source of recreation and Counter Strike: Global Offensive (CS: GO) is one of the top-listed online first-person shooting games. Numerous competitive games are arranged every year by Esports. Nonetheless, (i) No study has been conducted on video analysis and action recognition of CS: GO game-play which can play a substantial role in the gaming industry for prediction model (ii) No work has been done on the real-time application on the actions and results of a CS: GO match (iii) Game data of a match is usually available in the HLTV as a CSV formatted file however it does not have open access and HLTV tends to prevent users from taking data. This manuscript aims to develop a model for accurate prediction of 4 different actions and compare the performance among the five different transfer learning models with our self-developed deep neural network and identify the best-fitted model and also including major voting later on, which is qualified to provide real time prediction and the result of this model aids to the construction of the automated system of gathering and processing more data alongside solving the issue of collecting data from HLTV.

\end{abstract}


\section{Introduction}
Video games have become a significant part of the current century as a source of entertainment; thus, it owns a large portion of the entertainment market in the current world. This study focuses on a first-person shooting game named as Counter Strike: Global Offensive or shortly known ‘CS: GO’ where competition between two teams as attacking and defending roles takes place. While a match is taking place, different kinds of actions take place, kills, deaths, smoking, throwing a grenade, planting a bomb, getting injured, etc. can be shown as examples. The value of these parameters after a game varies from one player to another depending on their skills and experience. Reversely it can be said that, if the parameter values can be predicted accurately, not only the quality of a player in terms of skills and experience can be identified but also the winning and losing probability can also be determined even before the match have started. As the gaming market has grown substantially larger and still growing, as well as many huge tournaments and competitions are taking place all across the world, this kind of prediction that forecasts an individual’s and team’s attributes can be convenient and profitable for the related business industries.  \par
\begin{figure}[!tbp]
  \centering
  \begin{minipage}[b]{0.24\textwidth}
    \includegraphics[width=\textwidth]{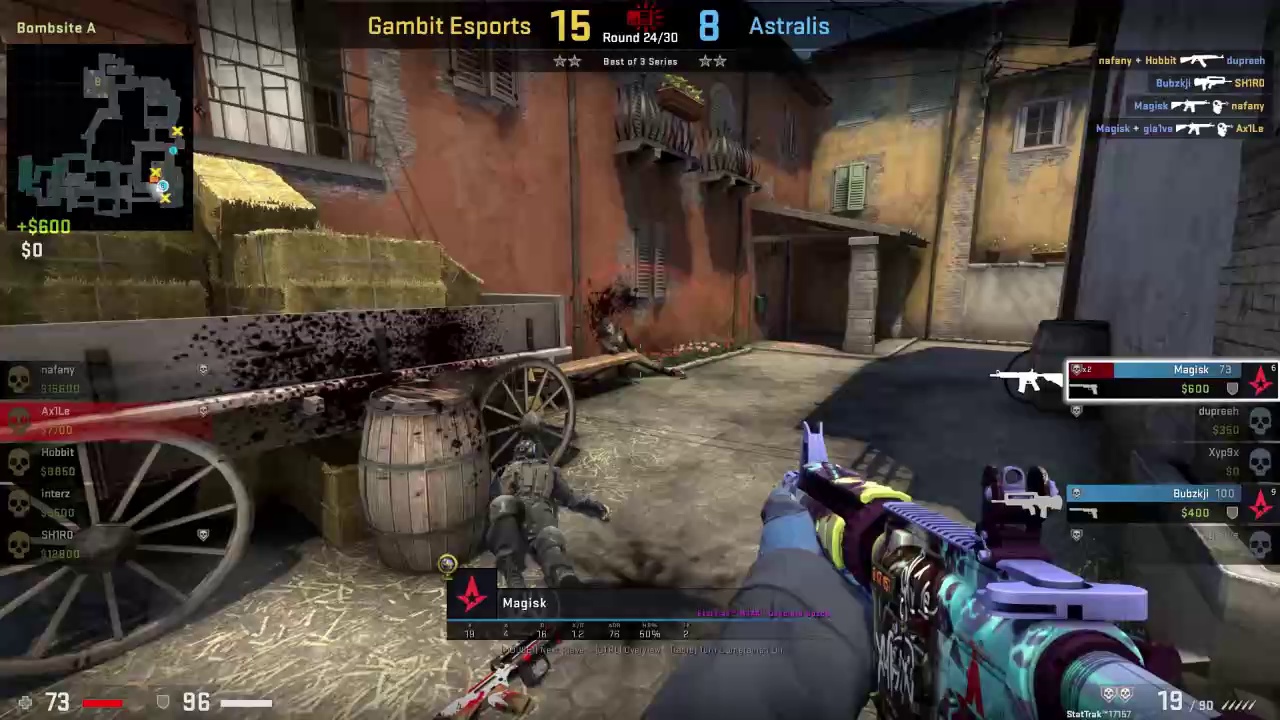}
    \caption{Kill.}
    \label{fig:x kill_frame}
  \end{minipage}
  \hfill
  \begin{minipage}[b]{0.24\textwidth}
    \includegraphics[width=\textwidth]{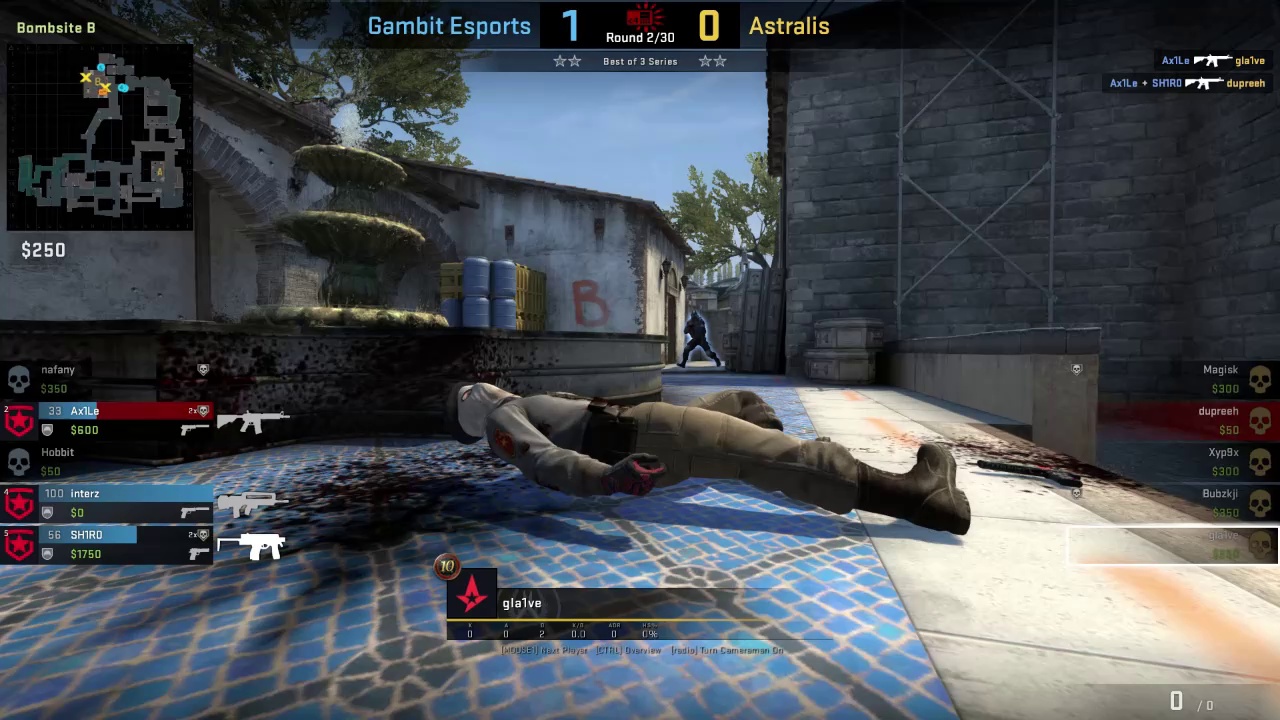}
    \caption{Death.}
    \label{fig:x death_frame}
  \end{minipage}
  \hfill

  \begin{minipage}[b]{0.24\textwidth}
    \includegraphics[width=\textwidth]{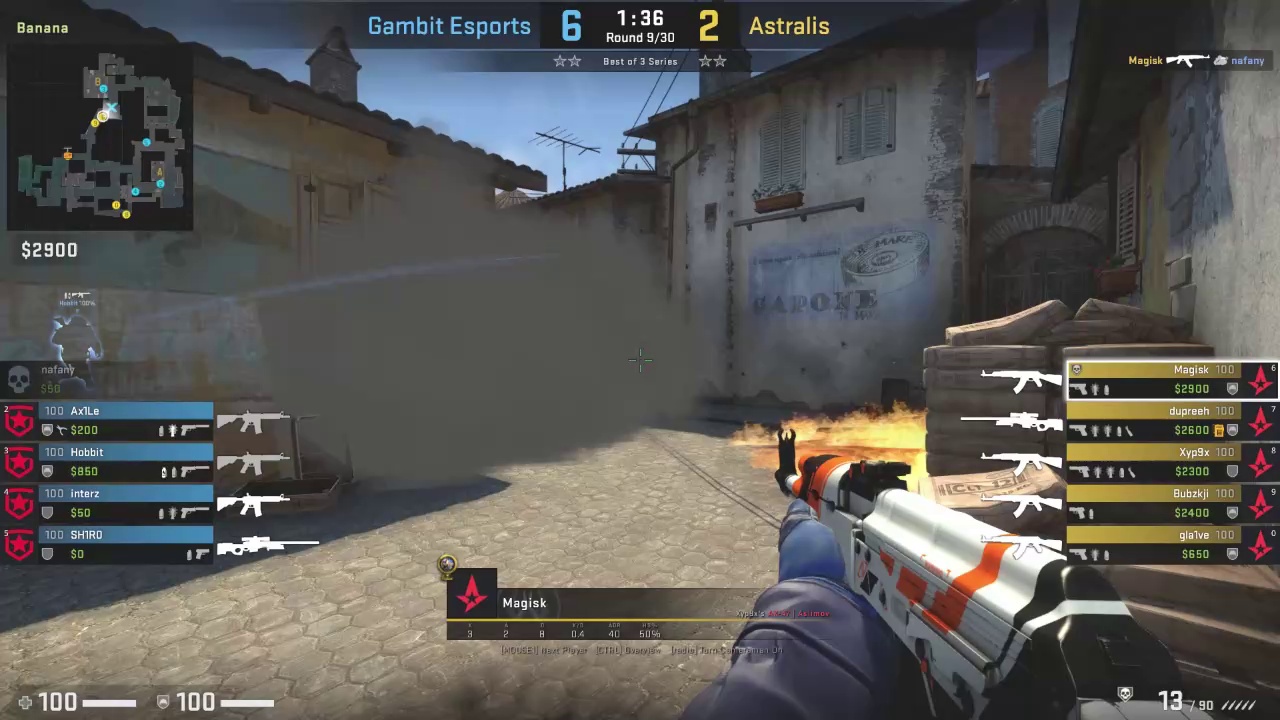}
    \caption{Smoke.}
    \label{fig:x smoke_frame}
  \end{minipage}
  \hfill
  \begin{minipage}[b]{0.24\textwidth}
    \includegraphics[width=\textwidth]{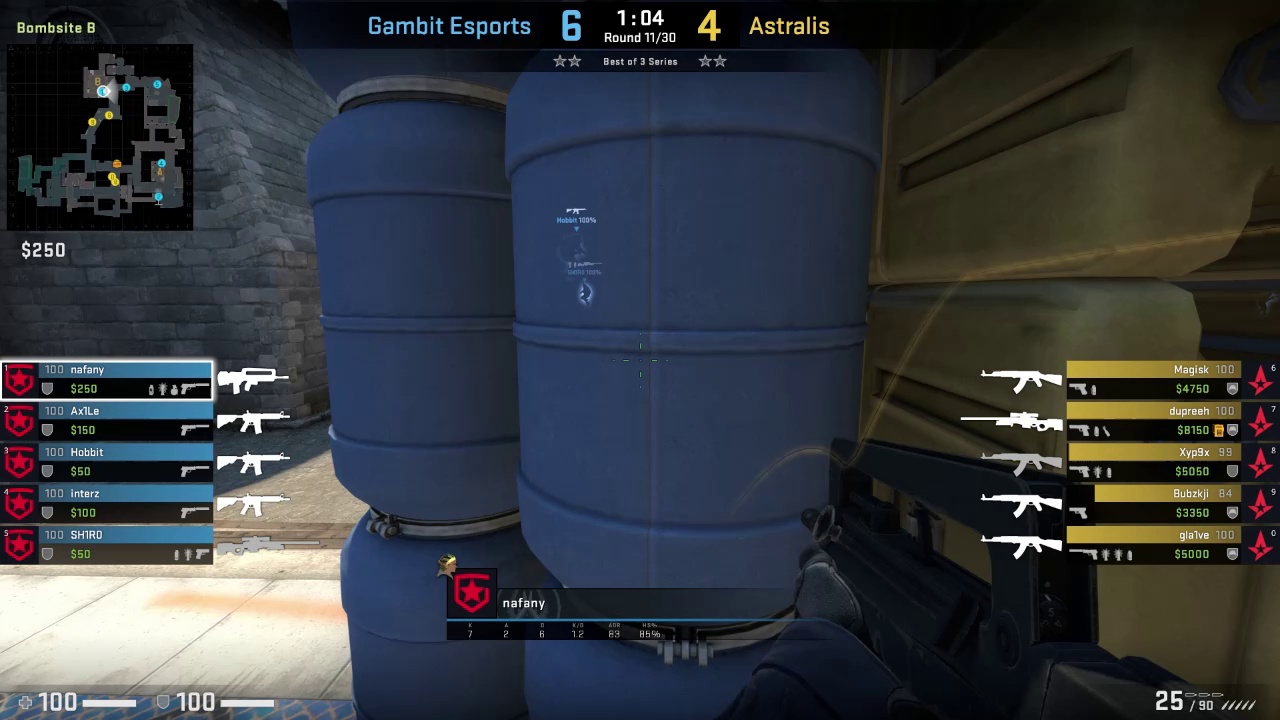}
    \caption{No Action.}
    \label{fig:x noaction_frame}
  \end{minipage}
\end{figure}

Our work is comprised of five different transfer learning models VGG16, Xception, Inception\_v3, Inception\_ResNet\_v2 and ResNet152\_v2 and majority voting as the focal point, analyzing them separately, finding out the best-fitted model for the prediction of four different actions of the CS: GO gameplay named as kill  [Fig \ref{fig:x kill_frame}], death [Fig \ref{fig:x death_frame}],smoke [Fig \ref{fig:x smoke_frame}] and no action [Fig \ref{fig:x noaction_frame}]. These classes were picked since they are the most frequent actions in the game and also have a higher value while developing a prediction model. During performing our study, we have faced various obstacles such as (i) Data collecting and pre-processing was challenging as HLTV provides only dem formatted files. (ii) No such study on real-time action detection for CS: GO  game has been conducted before (iii)Pre-processing of data for training purposes is tiresome and time-consuming. Particularly our contributions are listed below:
\par
\begin{itemize}

  \item Five different transfer learning models in addition to a self-developed deep neural network model have been used for video analysis. 

  \item  The model that has been developed in this study is capable of detecting actions in real time and provides a CSV formatted file for the prediction model simultaneously.
  
  \item  A system API has been constructed that allows the automated collection and processing of data which can aid to the later training period and enhance our system's performance. 
  
  \item Through our constructed API, the frequency of any of the mentioned actions occurring in a video can be determined.

\end{itemize}

In this manuscript, a short presentation of literature review on preceding action recognition studies has been introduced in the Section \ref{relatedwork}  followed by a brief description of our approach, model's and methods that have been used in the Section \ref{proposed}, which is comprised of three sub-sections. The discussion of the system model has been presented in \ref{system}, whereas \ref{data} talks about the data acquisition and preparation techniques and lastly \ref{model} illustrates a quick introduction of the used pre-trained model's. The performance metrics and result have been demonstrated in the Section \ref{performance} including confusion matrices for each of the models and finally, the study comes to an end with a conclusion part in Section \ref{conclusion}.

\section{Related Work} \label{relatedwork}
There are numerous studies have been conducted to detect action in real time using CNN, LSTM or various pre-trained models. However, no work has been conducted on CS: GO. In this section, we discuss the previously available papers that conducted work to detect various actions from video footage. \par

Jeyanthi el. al. built a model using the Inception\_ResNet\_v2 pre-trained model and Long Short Term Model (LSTM) for the purpose of human action recognition, where the pre-trained model is used to extract the features and the LSTM is used to learn the order of the action \cite{suresh2020inception}. Moreover, VGG16, ResNet152 and Inception\_v3 are also used as elements of comparison. In the end, the combination of LSTM and Inception\_ResNet\_v2 proves to be gaining the best accuracy, 92\% and 91\% on UCI 101 and HMDB 51 data sets respectively. Similar kind of work done by Yousry et. al. where he showed a three-fold technique which includes a pre-training stage where a generic dataset is used and then this pre-trained dataset is used to the aimed dataset  \cite{abdulazeem2021human}. Finally, LSTM is used combining five different model shapes. VGG16, Xception and DenseNet are used as the pre-trained models where VGG16 proves to be the most precise model with an accuracy of 93.48\%. \par
A different approach is illustrated by Bux Sargano et. al. for action recognition using a hybrid model of Support Vector Machines and K-Nearest Neighbor (SVM-KNN) classifier along with the pre-trained CNNs and excludes the use of any sequential CNNs \cite{sargano2017human} . His proposed merged SVM and KNN model achieves an accuracy of 98.15\%. Alexander et. al. suggested a strategy to put the good use of predictions from transform learning in action recognition into temporal localization \cite{iqbal2019enhancing}. By incorporating simple CNN with the pre-trained model can meet the challenges offered by two localization action datasets.\par
Tian et. al. developed a 3D convolutional neural network that can not only extract the spatial features but also the temporal features. Moreover, sub-data classification is integrated with the system to deal with the cases of insufficient datasets and this strategy was named as the internal transfer learning strategy \cite{wang2017internal}. This model attains 98.2\%  and 100\% accuracy on KTH and Weizmann datasets respectively while integrating with ITL. The use of 3D convolutional neural networks can also be observed in Noha et. al.’s study \cite{sarhan2020transfer}. More specifically Inflated 3D (I3D) has been implemented to recognize sign language. By incorporating optical flow images with the RGB images, higher accuracy is achieved on the ChaLearn249 Isolated Gesture Recognition dataset. \par
Unlike the previous studies, we offer an action detection and data collecting automation framework for CS: GO. In this study, our goal is to deal with CS: GO dem formatted data, ensure real-time counting of detected actions, and gather data using Deep neural network and various pre-trained models.

\section{Proposed CS: GO Real Time Action Recognition Method} \label{proposed}

From Section \ref{proposed} we get a clear insight of our proposed model which is divided into three parts. In part \ref{system} we talk about our system model and then we move on to part \ref{data} to discuss our data acquisition and pre-processing and finally in part \ref{model} we talk about our used pre-trained and our self-made deep neural network model.

\subsection{System Model} \label{system}
\begin{figure}[!t]
\centering
\includegraphics[scale=0.425]{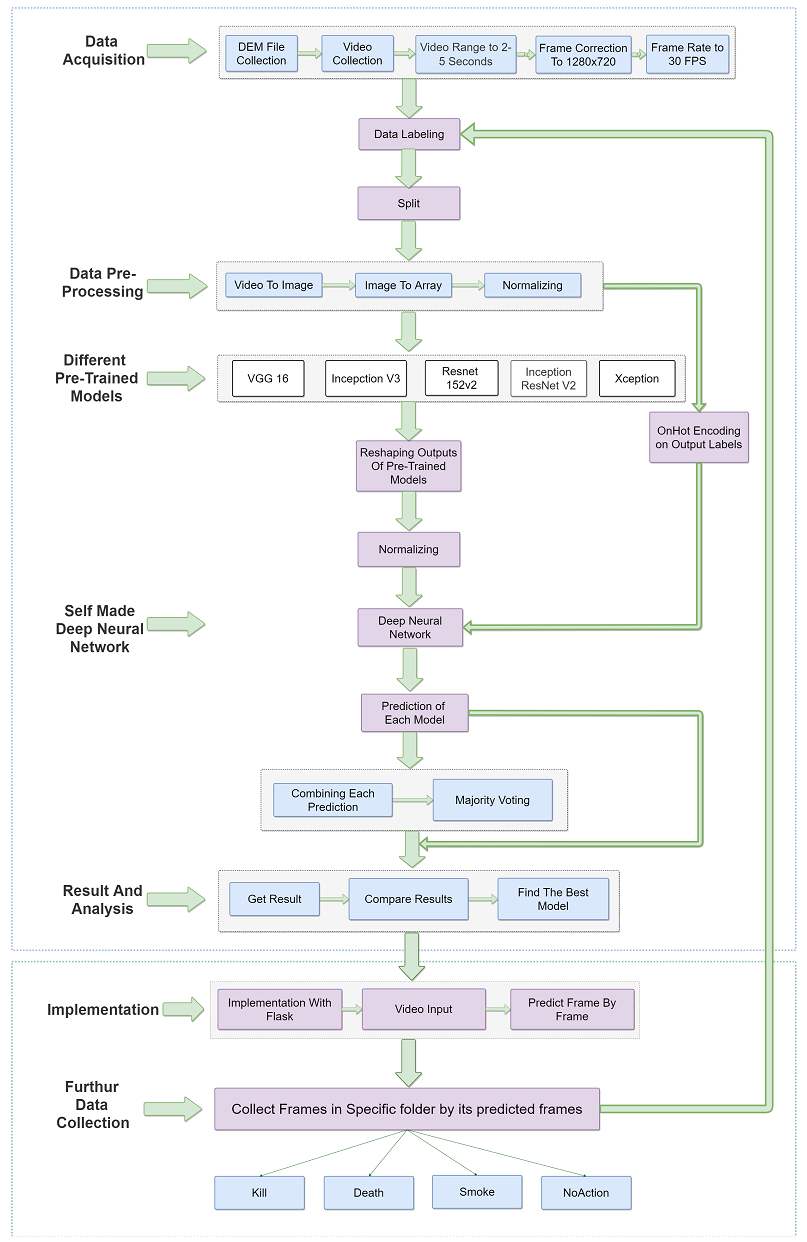}
\caption{System Model.}
\label{fig:x workflow}
\end{figure}

Our proposed system model (as depicted in Fig. \ref{fig:x workflow}), which allows us to detect various actions from CS: GO is real time is detailed in this section.\par

We concentrated on gaining data for four main categories: kill, death, smoke and no action which is basically everything else except the first three classes. All of these data were collected by DEM formatted files from one of the leading CS: GO public-site in the world HLTV. Later those DEM files were converted to videos. We made sure that all of the videos had a high-definition resolution of 1280 x 720 pixels and that all of the movies had the same frame rate of 30. Each clip lasted an average of 2 to 5 seconds, signifying the only time frame when the action took place. Secondly, we manually classified all of the videos, which was a time-consuming process but quite effective in assigning the correct labels. We retrieved several frames from each clip after separating our dataset into train, test, and validation segments. Later we transformed those frames into arrays and finally normalized the values that we got from the images. After we are done with the pre-processing part we passed those values through a number of pre-trained models which returned an array of values for each image and later to get better solutions again we normalized those values. We reshaped those data into 1-dimensional shapes to pass them into our model. We attempted to generate outputs that were identical to our one-hot encoded label outputs using our model. In our next step, we compared all of the pre-trained models' scores to find out which pre-trained model performed better. \par
We  have used VGG16, Xception, Inceptionv3, Inception\_ResNetv2 and ResNet152v2 as the transfer learning models to extract the detailed features from the input data and later developed a deep neural network model with five hidden layers to train the feature extracted data. In addition, we have performed majority voting on these five different approaches. Moreover, we have built an API after training our model which enabled us to predict random video frame by frame and later on the basis of our model's prediction we labeled those frames into different folders so that those frames later can be used to train our model more efficiently. In summary, we attempted to develop a system that allows us to combine existing solutions with our own to get superior outcomes. Fig \ref{fig:x workflow} represents our system model.

\subsection{Data Acquisition and Preparation} \label{data}

In this section we discuss our data collection and data pre-processing. \par

\begin{figure}[!t]
\centering
\includegraphics[scale=0.5]{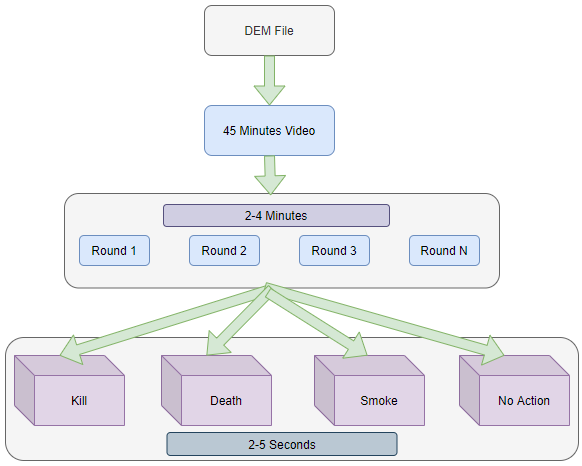}
\caption{Data Acquisition.}
\label{fig:x data}
\end{figure}

As stated before we collected DEM formatted files which is a game file and only playable by CS: GO game. Each DEM file is about 45 minutes long which contains 30 rounds of game-play. We manually recorded each DEM file with a Windows screen recorder. Later we cropped it into 30 different rounds to keep track of the rounds. And finally, we cropped individual actions with a duration of 2-5 seconds. Initially, we have collected more than 200 videos for each of the classes. Fig \ref{fig:x data} represents our workflow for data acquisition. Videos are just a stack of frames. The term "video" refers to the capture or transmission of moving visual images. Videos, like pictures, have width, height, and depth. We must specify at what speed images will be presented in a clip as video is a representation of visual images. Its frame rate can be defined in fps (frames per second). The number of pictures presented in a video every second is measured in frames per second (fps). Dealing with all frames of a clip might not be the best answer for our problem even though we're trying to anticipate a real-time event, and dealing with each frame could lead to a false hypothesis, and it's a time-consuming procedure. As a result, we divided the frame rate by each frame number and picked only those frames that resulted in the remainder value being set to zero. The entire procedure was carried out with the help of openCV3. Our original frame size was (1280x720). After performing some analysis, we discovered that most pre-trained models use the shape (224,224,3) as input. We transformed our image into this shape during loading frames. We normalized all of the picture arrays after converting them to an array in order to extract additional information from them. We normalized our input shape after extracting features before passing it into the Deep Neural Network model. As previously stated, we have identified four categories of activities in our research which are Kill, death, smoke and no action. Initially, we manually applied labels to each clip. We saved the label with the frame during converting video to picture by adding action to the frame name. To produce output categories, we used a single hot encoding. Based on the labels we previously established, we produced a separate output row for each label and filled it with binary 0 and 1 values.

\subsection{Model Specification} \label{model}
Within the part, we discuss various pre-trained models that we have used and our self-developed deep neural network model.

\subsubsection*{VGG-16 \cite{simonyan2014very}} \label{vgg}
\begin{figure}[!ht]
\centering
\includegraphics[scale=0.32]{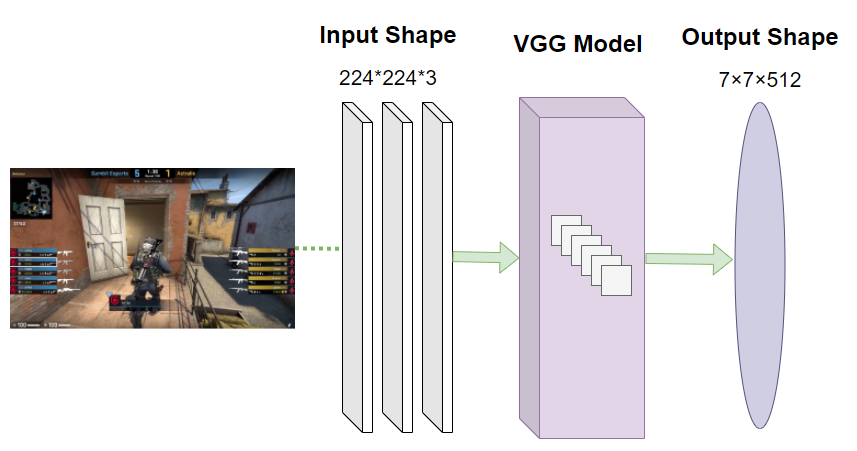}
\caption{Model Architecture with VGG-16.}
\label{fig:x vggarchi}
\end{figure}

It is one of the most commonly used pre-trained models which has more than 138 million parameters. Fig \ref{fig:x vggarchi} depicts the fundamental VGG-16 structure. We have used the basic architecture for VGG-16 and received an output shape of 7*7*512.

\subsubsection*{Inception V3 \cite{szegedy2016rethinking}}
\begin{figure}[!ht]
\centering
\includegraphics[scale=0.32]{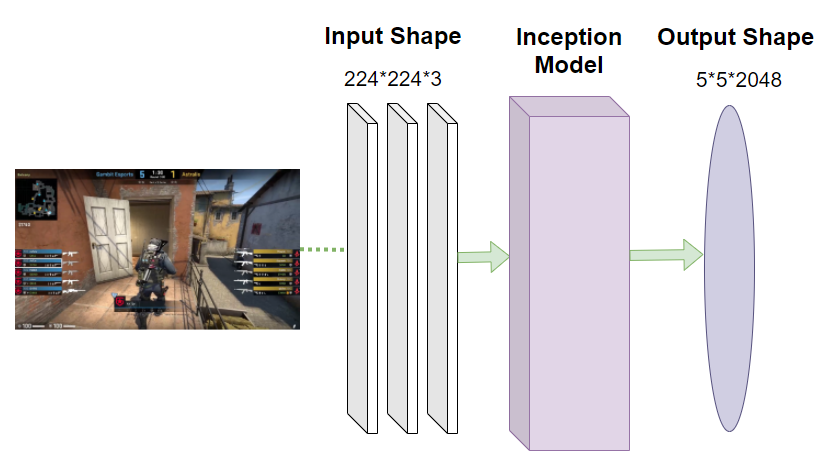}
\caption{Basic Model Architecture with Inception V3.}
\label{fig:x inceptionarchi}
\end{figure}

The inception model is also known as GoogLeNet. Fig \ref{fig:x inceptionarchi} shows the architecture of Inception V3. Utilizing inception v3 as our pre-trained model our output shape was 5*5*2048


\subsubsection*{Resnet 152v2 \cite{he2016deep}}
\begin{figure}[!ht]
\centering
\includegraphics[scale=0.32]{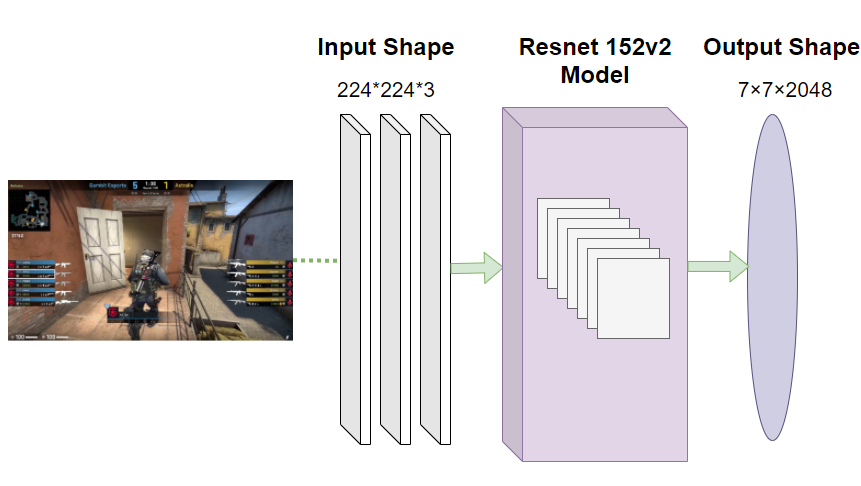}
\caption{Model Architecture With Resnet 152v2.}
\label{fig:x resnetarchi}
\end{figure}

Resnet 152v2 has over 60 million parameters. Fig \ref{fig:x resnetarchi} shows the architecture of Resnet 152v2. With the basic architecture of Resnet 152v2 and input shape as 224*244*3 we obtained our output shape as 7×7×2048. 

\subsubsection*{Inception ResNet V2\cite{inception}}
\begin{figure}[!ht]
\centering
\includegraphics[scale=0.3]{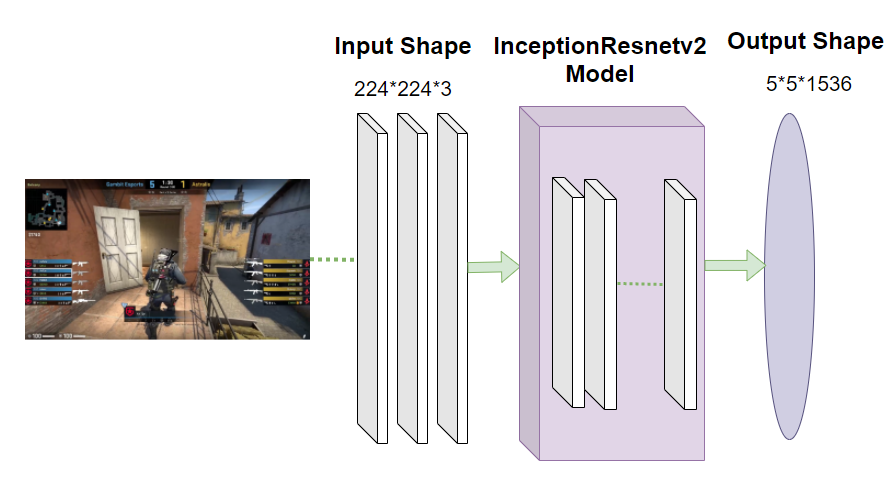}
\caption{Basic Model Architecture With Inception ResNet V2.}
\label{fig:x inceptionresnet}
\end{figure}

Inception ResNet V2 is a CNN model that is built based on the Inception architecture family. Here our output shape was 5*5*1536. Fig \ref{fig:x inceptionresnet} depicts the architecture of Inception Resnet 152v2.


\subsubsection*{Xception \cite{chollet2017xception}}
\begin{figure}[!ht]
\centering
\includegraphics[scale=0.3]{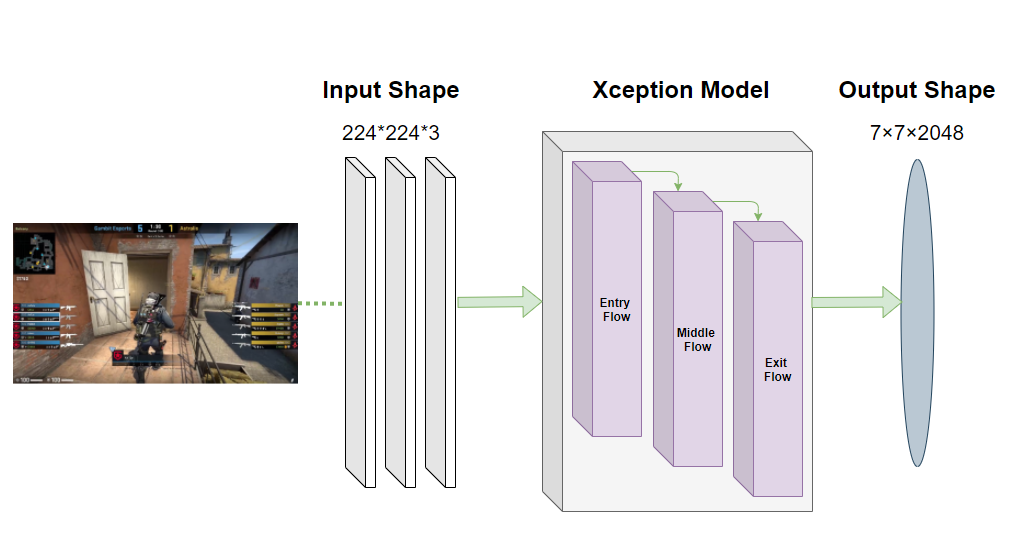}
\caption{Basic Model Architecture With Xception.}
\label{fig:x xception}
\end{figure}

Xception model is also built based on the inception model which is a deep convolutional neural network.
Fig \ref{fig:x xception} depicts the architecture of Xception in our system. 


\subsubsection*{Self Made Deep Neural Netwrok}
There are five fully connected layers in our Deep Neural Network model. This model's input layer uses the pre-trained model's normalized output value. We utilized four levels of dropout. We have set the dropout rate to 0.5, which ensures that 50 percent of the neurons in that layer are reduced at random every epoch. It means if a fully connected layer had 1024 neurons, and the dropout rate is set at 0.5 then only 512 neurons would be trained in the next fully connected layer. We used two different activation functions in our self-designed Deep Neural Network model. One of the activation functions is ReLU and the other one is softmax. In the output layer, we utilized the softmax activation function to transform actual value to probability. We got more accurate results by using the softmax activation algorithm. 
\subsection{API Implementation} \label{api_implementation}
\begin{figure}[!t]
\centering
\includegraphics[scale=0.5]{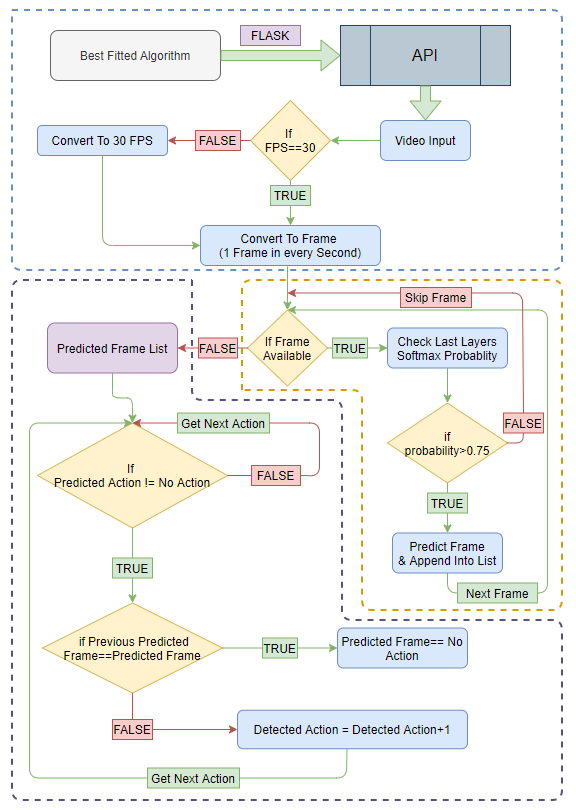}
\caption{API System Model.}
\label{fig:x api_workflow}
\end{figure}

Under this section we discuss about our API System which is divided into three main components depicted in fig \ref{fig:x api_workflow}. \par
We have implemented our best fitted algorithm with the aid of flask. Our API system is divided into three parts. In the first phase, we collect user input and then check to see if the FPS is 30. If our input data is not 30 frames per second, we convert to 30 FPS to ensure that we obtain one frame per second and then finally transform our video data into a sequence of frames. We cycle through all of the frames in the following component. As previously stated, in the final layer, which calculates the probability of each action, we utilized softmax as our activation function. We only keep the values in a list, if the probability of each frame's predicted action is more than 0.75. The rest of the frames are omitted. This probability checking ensure us that we can only have flawless predicted frames. In our final part, we cycle through the predicted frame list ignoring all No Action values. Then we check if our previous predicted value is the same as the predicted value and change the predicted value to No Action if our condition is met. Otherwise, we increment our count for the predicted action. The purpose of this further checking is to ensure that we do not count the same action twice since actions might occasionally last longer than one second, which can be problematic when we're predicting each frame.

\section{Performance Evaluation} \label{performance}
In this section, we analyze our overall performance which is divided into two parts. In part \ref{system_performance} we compare and evaluate our proposed model's performance and in part \ref{api_performance} we talk about our API's performance.
\subsection{System Performance Evaluation} \label{system_performance}
\begin{figure}[!t]
\centering
\includegraphics[scale=0.45]{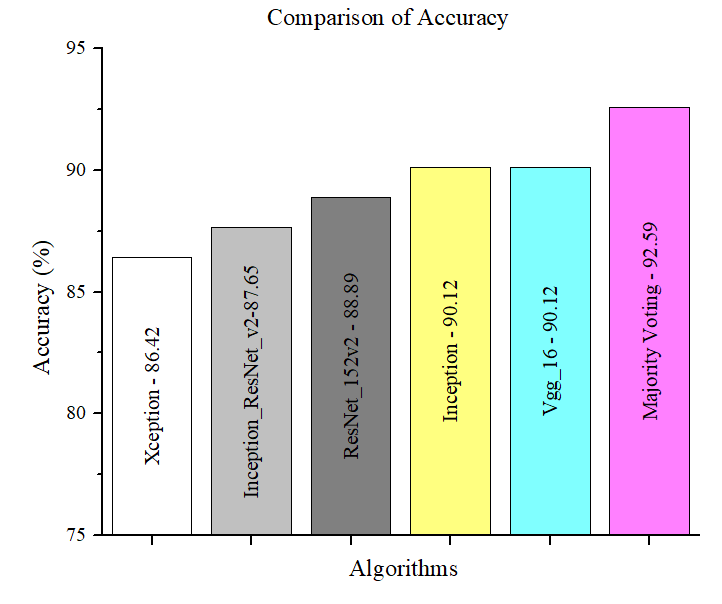}
\caption{Majority Voting : System Accuracy.}
\label{fig:x accuracy_graph}
\end{figure}

As mentioned earlier, ﬁve distinct pre-trained image classiﬁcation models have been applied for feature extraction and a self-developed deep neural network has been implemented for training and later by combining all the predictions of the models we performed majority voting. Among our implemented individual models for extraction VGG-16 and Inception\_v3 performs better than the rest by providing 90.12\% accuracy for both cases. ResNet152\_v2, Inception\_ResNet\_v2 and Xception have relatively less accuracy, but neither of them performed considerably poorly on the dataset and their accuracy value of 88.89\%, 87.65\% and 86.42\% respectively infers the precedent statement. Moreover, the later inclusion of majority voting exceeds the peak accuracy value of the individual pre-trained models and provides more precise prediction with 92.59\% accuracy. Fig. \ref{fig:x accuracy_graph} depicts the comparison of the described accuracy value among the different classifiers.

\begin{table}[!t]
\caption{Model Performance.}
\begin{center}

\begin{tabular}{|c|c|c|c|c|c| } 
\hline
Model & Action & Precision & F-1 Score & Sensitivity & Accuracy\\
\hline \hline
 & Kill &  0.8421  & 0.8205 & 0.80 & \\ 
{Xception}& Death & 1.00  & 0.8889  & 0.80 & {86.42}\\ 
& Smoke & 0.8695  &  0.9091 & 0.9523 &\\
& No Action & 0.7826  & 0.8372 & 0.90 &\\ 
\hline
Inception & Kill & 1.00  & 0.8235 & 0.70 & \\
\_Resnet & Death & 0.9047  & 0.9268 & 0.95 & {87.65}\\ 
V-2 & Smoke & 0.8695  & 0.9091 & 0.9523 &\\ 
& No Action & 0.7826  & 0.8372 & 0.90 &\\ 
\hline
& Kill & 0.8421  & 0.8205 & 0.80 & \\ 
ResNet & Death & 0.95  & 0.95 & 0.95 &  {88.89}\\ 
152v2 & Smoke & 0.9473  & 0.90 & 0.8571 &\\ 
& No Action & 0.8261  & 0.8837 & 0.95 &\\ 
\hline
& Kill & 0.89  & 0.87 & 0.85 & \\ 
Inception & Death & 0.90  & 0.93 & 0.95 & 90.12\\ 
V-3 & Smoke & 0.87  & 0.91 & 0.95 &\\
& No Action & 0.94  & 0.89 & 0.85 &\\ 
\hline
& Kill & 0.7826  & 0.8372 & 0.90 & \\ 
VGG-16 & Death & 0.95  & 0.95 & 0.95 & {90.12}\\ 
& Smoke & 0.95  & 0.9268 & 0.9047 &\\ 
& No Action & 0.9444  & 0.8947 & 0.85 &\\ 
\hline
& Kill & 1.00  & 0.9744 & 0.95 & \\
Majority & Death & 0.9474  & 0.9231 & 0.9 & {92.59}\\ 
Voting & Smoke & 0.9048  & 0.9268 &0.95 &\\ 
& No Action & 0.9545  & 0.9767 & 1.00 &\\ 
\hline

\end{tabular}
\label{modelscore}
\end{center}

\end{table}

\begin{figure}[!t]
\centering
\includegraphics[scale=0.24]{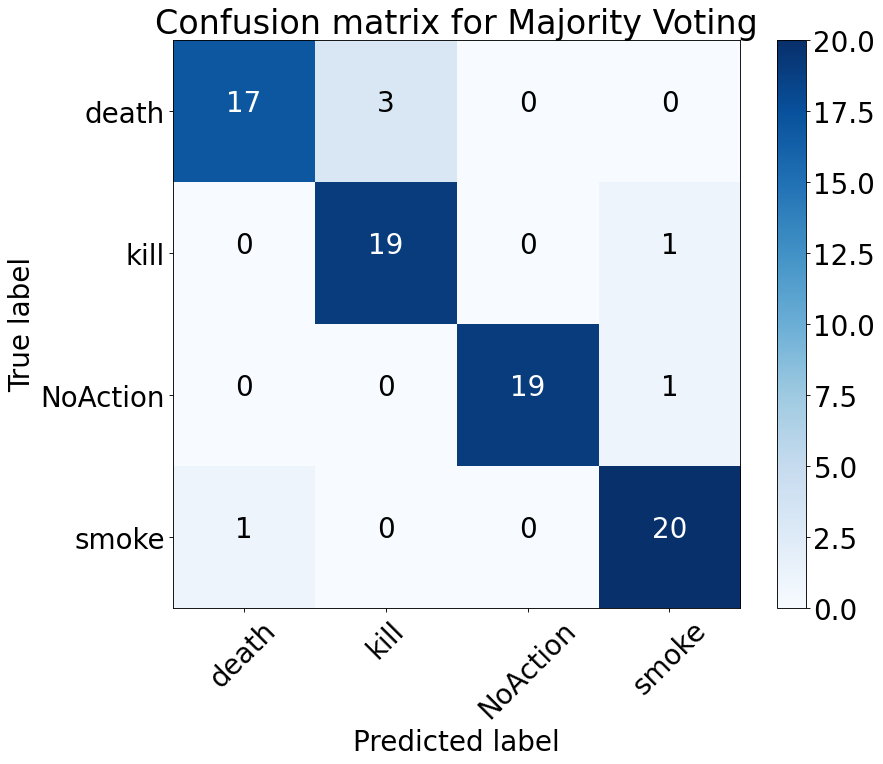}
\caption{Majority Voting : Confusion Matrix.}
\label{fig:x Confusion}
\end{figure}

Table I depicts the overall model performance with these accuracy values along with precision, f1 score and sensitivity values for each of the four actions of every pre-trained model. As illustrated from the table, on average the action ‘death’ achieves the finest performance values for every parameter and 1.00 in precision for Xception as well as 0.95 in all the parameter for VGG16 are some of the examples of that best performance. In addition, Fig \ref{fig:x Confusion} illustrates the confusion matrix of majority voting for the four classes. As it can be seen that, the detection of smoke through majority voting is the most accurate as all the 20 out of 20 data example of smoke is detected accurately.

\subsection{API Performance Evaluation} \label{api_performance}

\begin{figure}[!ht]
\centering
\includegraphics[scale=0.5]{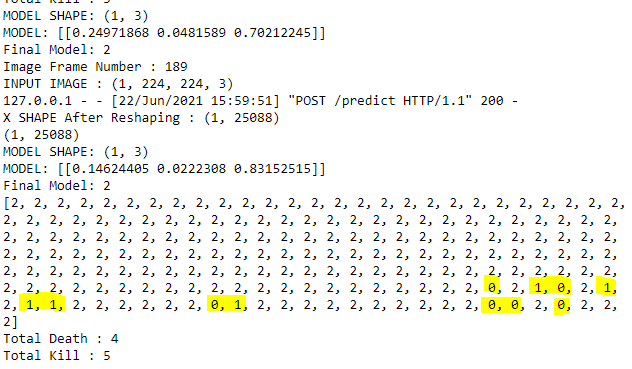}
\caption{API Performance Evaluation.}
\label{fig:x API_result}
\end{figure}

Overall performance of our API was exceptional. We are able to correctly count the number of times an action has occurred using our API. As can be seen in Fig \ref{fig:x API_result} our API successfully counts the number of actions in an input video that comprised four deaths and five kills. Here, 0 represents kill, 1 represents death, 2 represents no action, and 3 represents smoke. As there is no smoke in this video, none of the frames are identified as smoke. Initially, 0 appears six times and 1 appears five times, implying that six frames are predicted as kills and five frames are predicted as deaths which means kill count is six and death count is five. However, using our algorithm, this problem is resolved and our API can provide us with the proper count.

\section{Conclusion} \label{conclusion}
As we can see from the performance evaluation section, the highest accuracy is gained by majority voting (92.59\%), which infers the developed model as quite precise and reliable to predict the 4 different actions. Through our system we will be able to detect various actions from 45 minutes competitive game-play of CS: GO game and upon detecting actions our system will generate data in real time in CSV format. Later these data can be used for a prediction model which is our one of the end goals of this study. Moreover, through API our system will be able to automate the data collection process that will improve the accuracy and other performance metrics after training on the increased data set. As mentioned in the manuscript, five transfer learning models have been used in the research, whereas there are many more transfer learning models with their own distinctive characteristics. In future, we are also expecting to work with these models and achieve variation in the performance metrics.

\vspace{12pt}

\end{document}